\pdfoutput=1

\documentclass[a4paper]{llncs}
\usepackage{makeidx}  
\usepackage{amssymb}
\usepackage{listings}
\usepackage{color}
\usepackage[pdftex]{graphicx}
\usepackage{array}
\usepackage{url}
\usepackage{wrapfig}
\newcommand{\keywords}[1]{\par\addvspace\baselineskip
\noindent\keywordname\enspace\ignorespaces#1}
\usepackage{float}
\usepackage{algorithm,algpseudocode}

\usepackage{subfig}

\usepackage{stfloats}
\usepackage{listings}
\usepackage{color}

\definecolor{dkgreen}{rgb}{0,0.6,0}
\definecolor{gray}{rgb}{0.5,0.5,0.5}
\definecolor{mauve}{rgb}{0.58,0,0.82}

\begin{document}
%
\title{Image Captioning using Deep Neural Architectures}

\author{Parth Shah (orcid.org/0000-0002-7880-2228) \and Vishvajit Bakarola (orcid.org/0000-0002-6602-9194) \and Supriya Pati (orcid.org/0000-0001-5528-6104) }

\institute{Chhotubhai Gopalbhai Patel Institute of Technology, Bardoli, India\\\email{parthpunita@yahoo.in}\\ \email{vishvajit.bakrola@utu.ac.in} \\ \email{supriya.pati@utu.ac.in}}


%


\maketitle

\begin{abstract}
Automatically creating the description of an image using any natural languages sentence like English is a very challenging task. It requires expertise of both image processing as well as natural language processing. This paper discuss about different available models for image captioning task. We have also discussed about how the advancement in the task of object recognition and machine translation has greatly improved the performance of image captioning model in recent years. In addition to that we have discussed how this model can be implemented. In the end, we have also evaluated the performance of model using standard evaluation matrices.
\end{abstract}


%

\section{Introduction}

A single image can contain large amount of information in it. Humans have ability to parse this large amount of information by single glance of it. Humans normally communicate though written or spoken language. They can use languages for describing any image. Every individual will generate different caption for same image. If we can achieve same task with machine it will be greatly helpful for variety of tasks. However, generating captions for an image is very challenging task for machine. In order to perform caption generation task by machine, it requires brief understanding of natural language processing and ability to identify and relate objects in an image. Some of the early approaches that tried to solve these challenge are often based on hard-coded features and well defined syntax. This limits the type of sentence that can be generated by any given model. In order to overcome this limitation the main challenge is to make model free of any hard-coded feature or sentence templates. Rule for forming models should be learned from the training data.

Another challenge is that there are large number of images available with their associated text available in the ever expanding internet. However, most of them are noisy hence it can not be directly used in image captioning model. Training an image captioning model requires huge dataset with properly available annotated image by multiple persons.

In this paper, we have studied collections of different existing natural image captioning models and how they compose new caption for unseen images. We have also presented results of our implementation of these model and compared them.

Section 2 of this paper describes Related Work in detail. Show \& Tell model in detailed is described in Section 3. Section 4 contains details about implementation environment and dataset. Results and Discussion is provided in detail in Section 5. At the end we provided our concluding remarks in section 6.




\section{Related work}
Creating captioning system that accurately generate captions like human depends on the connection between importance of object in image and how they will be related to other objects in image. Image can be described using more than one sentence but to efficiently train the image captioning model we requires only single sentence that can be provided as a caption. This leads to problem of text summarization in natural language processing.

There are mainly two different way to perform the task of image captioning. These two types are basically retrieval based method and generative method. From that most of work is done based on retrieval based method. One of the best model of retrieval based method is Im2Txt model \cite{ordonez2011im2text}. It was proposed by Vicente Ordonez, Girish Kulkarni and Tamara L Berg. Their system is divided into mainly two part 1) Image matching and 2) Caption generation.
First we will provide our input image to model. Matching image will be retrieved from database containing images and its appropriate caption. Once we find matching images we will compare extracted high level objects from original image and matching images. Images will then reranked based on the content matched. Once it is reranked caption of top-n ranked images will be returned. The main limitation of these retrieval based method is that it can only produce captions which are already present in database. It can not generate novel captions.

This limitation of retrieval based method is solved in generative models. Using generative models we can create novel sentences. Generative models can be of two types either pipeline based model or end to end model. Pipeline type models uses two separate learning process, one for language modeling and and one for image recognition. They first identify objects in image and provides the result of it to language modeling task. While in end-to-end models we combine both language modeling and image recognition models in single end to end model \cite{karpathy2015deep}. Both part of model learn at the same time in end-to-end system. They are typically created by combination of convolutional and recurrent neural networks.

Show \& Tell model proposed by Vinyals et al. is of generative type end-to-end model. Show \& Tell model uses recent advancement in image recognition and neural machine translation for image captioning task. It uses combination of Inception-v3 model and LSTM cells \cite{7505636}.
Here Inception-v3 model will provides object recognition capability while LSTM cell provides it language modeling capability \cite{szegedy2015going}\cite{szegedy2015rethinking}.

\section{Show \& Tell Model}
Recurrent neural networks generally used in neural machine translation \cite{rnnintro}. They encodes the variable length inputs into a fixed dimensional vectors. Then it uses these vector representation to decode to the desired output sequence \cite{DBLP:journals/corr/WuSCLNMKCGMKSJL16}\cite{sutskever2014sequence}. Instead of using text as input to encoder Show \& Tell model uses image as input. This image is then converted to word vector and then this word vector is translated to caption using Recurrent neural networks as decoder.

To achieve this goal, Show \& Tell model is created by hybridizing two different models. It takes input as the image and provides it to Inception-v3 model. At the end of Inception-v3 model single fully connected layer is added. This layer will transform output of Inception-v3 model into word embedding vector. We input this word embedding vector into series of LSTM cell. LSTM cell provides ability to store and retrieve sequential information through time. This helps to generate the sentences with keeping previous words in context.
\begin{figure}[h]
  \centering
  \includegraphics[width=8cm]{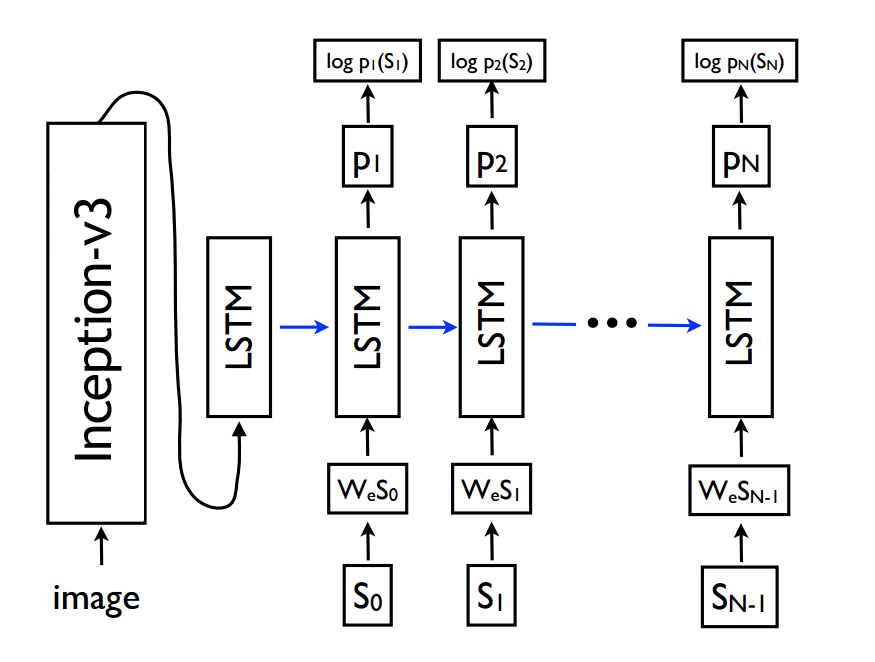}
  \caption{Architecture of Show \& Tell Model}\label{show_and_tell}
\end{figure}
Training of Show \& Tell model can be divided into two part. First part is of training process where model learns its parameters. While second part is of testing process. In testing process we infer the captions and we compare and evaluate these machine generated caption with human generated captions.
\subsection{Training}
During training phase we provides pair of input image and its appropriate caption to Show \& Tell model. Inception-v3 part of model is trained to identify all possible objects in an image. While LSTM part of model is trained to predict every word in the sentence after it has seen image as well as all previous words. For any given caption we add two additional symbols as start word and stop word. Whenever stop word is encountered it stop generating sentence and it marks end of string. Loss function for model is calculated as
\begin{equation}
L(I, S) = - \sum_{t=1}^N \log p_t(S_t) \; .
\end{equation}
where ${I}$ represent input image and ${S}$ represent generated caption. ${N}$ is length of generated sentence. ${p_t}$ and ${S_t}$ represent probability and predicted word at the time ${t}$ respectively. During the process of training we will try to minimize this loss function.

\subsection{Inference}

From various approaches to generate caption a sentence from given image Show \& Tell model uses Beam Search to find suitable words to generate caption. If we keep beam size as K, it recursively consider K best word at each output of the word. At each step it will calculate joint probability of word with all previously generated word in sequence. It will keep producing the output until end of sentence marker is predicted. It will select sentence with best probability and outputs it as caption.

\section{Implementation}
For evaluation of image captioning model we have implemented Show \& Tell model. Details about dataset,implementation tool and implementation environment is given as follows:

\subsection{Datasets}

For task of image captioning there are several annotated images dataset are available. Most common of them are Pascal VOC dataset and MSCOCO Dataset. In this work MSCOCO image captioning dataset is used. MSCOCO is a dataset developed by Microsoft with the goal of achieving the state-of-the-art in object recognition and captioning task. This dataset contains collection of day-to-day activity with theri related captions. First each object in image is labeled and after that description is added based on objects in an image. MSCOCO dataset contains image of around 91 objects types that can be easily recognizable by even a 4 year old kid. It contains around 2.5 million objects in 328K images. Dataset is created by using crowdsourcing by thousonds of humans \cite{lin2014microsoft}.

\subsection{Implementation tool and environment}
For the implementation of this experiment we have used machine with Intel Xeon E3 processor with 12 cores and 32GB RAM running CentOS 7. Tensorflow liberary is used for creating and training deep neural networks. Tensorflow is a deep learning library developed by Google\cite{abadi2016tensorflow}. It provides heterogeneous platform for execution of algorithms i.e. it can be run on low power devices like mobile as well as large scale distributed system containing thousands of GPUs. To define structure of our network tensorflow uses graph definition. Once graph is defined it can be executed on any supported devices.

\section{Results and Discussion}
\subsection{Results}
By the implementation of the Show \& Tell model we can able to generate moderately comparable captions with compared to human generated captions.
First of all it model will identify all possible objects in image.
\begin{figure}[h]
  \centering
  \includegraphics[width=8cm,height=8cm]{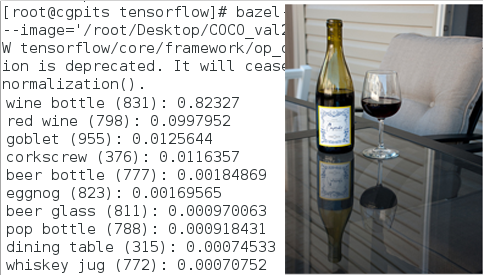}
  \caption{Generated word vector from Sample Image}\label{word_vec}
\end{figure}

\begin{figure}[h]
  \centering
  \includegraphics[width=9cm,height=4cm]{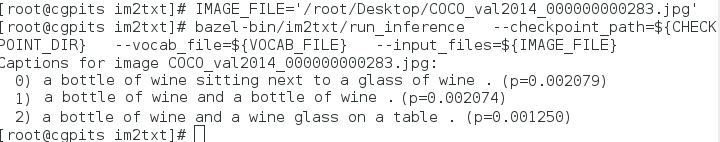}
  \caption{Generated caption from word vector for Sample Image}\label{result4}
\end{figure}

As shown in Fig. \ref{word_vec} Inception-v3 model will assign probability of all possible object in image and convert image into word vector. This word vector is provided as input to LSTM cells which will then form sentence from this word vector as shown in Fig. \ref{result4} using beam search as described in previous section.

\subsection{Evaluation Matrices}
To evaluating of any model that generate natural language sentence BLEU (Bilingual Evaluation Understudy) Score is used. It describes how natural sentence is compared to human generated sentence \cite{papineni2002bleu}. It is widely used to evaluate performance of Machine translation. Sentences are compared based on modified n-gram precision method for generating BLEU score \cite{jurafsky2000speech}. Where precision is calculated using following equation:
\begin{equation}\label{bleu}
  p_n=\frac{\sum_{C\in\{Candidates\}}\sum_{ngram\in C}Count_{clip}(ngram)}{\sum_{C'\in\{Candidates\}}\sum_{ngram'\in C'}Count(ngram')}
\end{equation}

To evaluate our model we have used image from validation dataset of MSCOCO Dataset. Some of captions generated by Show \& Tell model is shown as follows:
\begin{figure}[h]
\subfloat[Input image]{
\includegraphics[width=4cm,height=5cm]{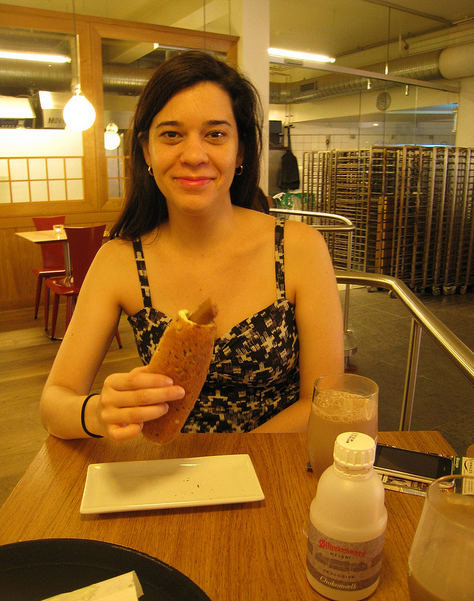}}\subfloat[Generated caption]{\includegraphics[width=9cm,height=5cm]{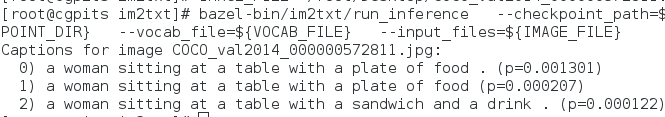}
}
\caption{Experiment Result}\label{sample4}
\end{figure}

As you can see in Fig. \ref{sample4}, generated sentence is ``a woman sitting at a table with a plate of food.", while actual human generated sentence are ``The young woman is seated at the table for lunch, holding a hotdog.", ``a woman is eatting a hotdog at a wooden table.", ``there is a woman holding food at a table.", ``a young woman holding a sandwich at a table." and ``a woman that is sitting down holding a hotdog.". This result in BLEU score of 63 for this image.
\begin{figure}[h]
\subfloat[Input image]{
\includegraphics[width=4cm,height=5cm]{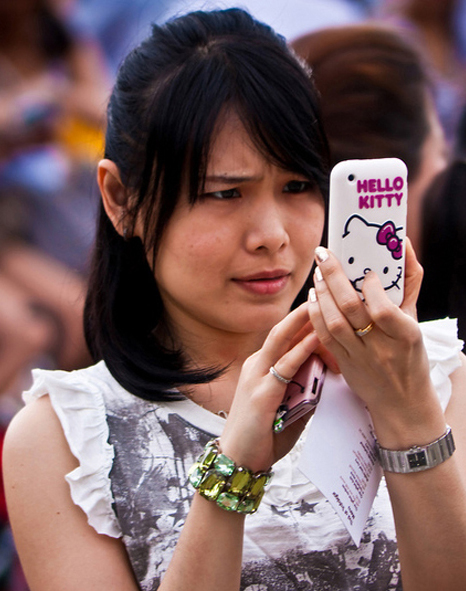}}\subfloat[Generated caption]{\includegraphics[width=9cm,height=5cm]{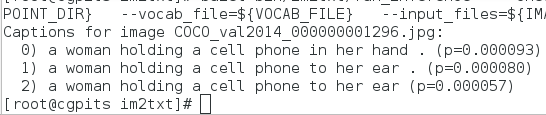}
}
\caption{Experiment Result}\label{sample8}
\end{figure}

Similarly in Fig. \ref{sample8}, generated sentence is ``a woman holding a cell phone in her hand." while actual human generated sentence are ``a woman holding a Hello Kitty phone on her hands", ``a woman holds up her phone in front of her face", ``a woman in white shirt holding up a cellphone", ``a woman checking her cell phone with a hello kitty case" and ``the asian girl is holding her miss kitty phone". This result in BLEU score of 77 for this image.

While calculating BLEU score of all image in validation dataset we get average score of 65.5. Which shows that our generated sentence are very similar compared to human generated sentence.

\section{Conclusion}
We can conclude from our findings that we can combine recent advancement in Image Labeling and Automatic Machine Translation into an end-to-end hybrid neural network system. This system is capable to autonomously view an image and generate a reasonable description in natural language with better accuracy and naturalness.




\bibliographystyle{ieeetr}
\bibliography{dl_image_caption.bib}
%

%
%

\end{document}